\documentclass[APS,Times2COL]{WileyNJDv5} %STIX1COL,STIX2COL,STIXSMALL

\articletype{ORIGINAL RESEARCH}%

\received{Date Month Year}
\revised{Date Month Year}
\accepted{Date Month Year}
\journal{IET Intell. Transp. Sys.}
\journal{Page}
\volume{00}
\copyyear{2024}
\startpage{1}

\begin{document}

\title{Evaluating Driver Readiness in Conditionally Automated Vehicles from Eye-Tracking Data and Head Pose}

\author[1]{Mostafa Kazemi}
\author[2]{Mahdi Rezaei}
\author[3]{Mohsen Azarmi}

\authormark{KAZEMI \textsc{et al.}}
%Short Title
\titlemark{Evaluating Driver Readiness in Conditionally Automated Vehicles}

\address[]{\orgdiv{Institute for Transport Studies}, \orgname{University of Leeds}, \orgaddress{\state{Leeds}, \country{UK}}}

%\address[2]{\orgdiv{Department Name}, \orgname{Institution Name}, \orgaddress{\state{State Name}, \country{Country Name}}}

%\address[3]{\orgdiv{Department Name}, \orgname{Institution Name}, \orgaddress{\state{State Name}, \country{Country Name}}}

\corres{Associate Professor Mahdi Rezaei\\ Institute for Transport Studies, University of Leeds, Leeds, LS2 9JT, UK.\email{m.rezaei@leeds.ac.uk}}

%\presentaddress{This is sample for present address text this is sample for present address text.}

%\fundingInfo{Text}
%\JELinfo{ejlje}

\abstract[Abstract]
{As automated driving technology advances, the role of the driver to resume control of the vehicle in conditionally automated vehicles becomes increasingly critical. In the SAE Level 3 or partly automated vehicles, the driver needs to be available and ready to intervene when necessary. This makes it essential to evaluate their readiness accurately. This article presents a comprehensive analysis of driver readiness assessment by combining head pose features and eye-tracking data. The study explores the effectiveness of predictive models in evaluating driver readiness, addressing the challenges of dataset limitations and limited ground truth labels. Machine learning techniques, including LSTM architectures, are utilised to model driver readiness based on the Spatio-temporal status of the driver's head pose and eye gaze. The experiments in this article revealed that a Bidirectional LSTM architecture, combining both feature sets, achieves a mean absolute error of 0.363 on the DMD dataset, demonstrating superior performance in assessing driver readiness. The modular architecture of the proposed model also allows the integration of additional driver-specific features, such as steering wheel activity, enhancing its adaptability and real-world applicability.}

\keywords{Autonomous vehicles, Driver readiness, Head Pose Estimation, Gaze estimation, Eye-tracking}

%\jnlcitation{\cname{%
%\author{Kazemi, M.}
%\author{Rezaei, M}} and
%\author{Azarmi, M.}.
%\ctitle{Evaluating Driver Readiness in Conditionally Automated Vehicles from Eye-Tracking Data and Head Pose.} \cjournal{\it IET Intell. Transp. Syst.} \cvol{2023;00(00):1--18}.}

\maketitle

\section{INTRODUCTION}\label{sec1}

According to data published by the World Health Organisation (WHO) \cite{WHO}, over 1.35 million people lose their lives annually due to road-related accidents, and millions more suffer severe injuries. Within these unfortunate events, it has been observed that 90\% of all accidents are attributed to human errors \cite{amrun2023review}. The development of technology has given rise to the hope that the concept of automated driving can address these challenges \cite{boggs2020exploratory}. Removing the driver from the driving process and transferring control of the vehicle to an automated driving function can be a step towards eliminating human error in accidents \cite{boelhouwer2019should}.

In L3 conditionally automated driving, the system temporarily takes the responsibility for steering, acceleration, and braking, along with the vehicle-related duties, while the driver has the option to engage in Non-Driving Related Tasks (NDRTs) like reading the news, watching a movie, or relaxing in the driver's seat \cite{braunagel2017ready}. However, the human drivers must be prepared to regain control in cases of emergencies, like when the system malfunctions and prompts a Take-Over Request (TOR) or when the situation goes beyond the system's capabilities \cite{morales2020automated}. During such instances, the ability to perceive and understand road and traffic conditions, essentially situational awareness, becomes crucial in determining immediate actions. At this stage, the question arises whether the driver is ready to take back control of the vehicle.

Prior to initiating a TOR, it is crucial to gather both internal and external data regarding the driver's condition and traffic context \cite{rezaei2023} to ensure a secure outcome \cite{du2020predicting}. This is especially important when dealing with take-over procedures necessitated by sudden emergencies. Thus, the development of Driver Monitoring Systems (DMS) becomes imperative \cite{dogan2022evaluation}. These systems should possess the capability to assess the driver's physical and mental condition and subsequently correlate this data with the specific road scenario \cite{du2020predicting}. To achieve this objective, the DMS accumulates baseline metrics for various driver state indicators, including factors like the driver's gaze, posture, and physiological measurements \cite{kang2013various}. These baseline metrics are then employed to compute a measure of deviation, enabling the system to evaluate the extent of the driver's readiness.

In this paper, we investigated the strengths and weaknesses of the existing models and datasets for assessing driver readiness in conditionally automated vehicles. Building on this understanding, 
%we focus on examining the driver's facial expressions inside the vehicle. Our 
we develop a non-intrusive computer vision-based framework for evaluating driver readiness using facial landmark features. The proposed model tracks facial features and analyses their geometric configuration as features to estimate the driver's head pose and gaze directions. 

%We employ state-of-the-art Convolutional Neural Network models in computer vision to improve the accuracy of feature extraction from eye-tracking data and head pose as reliable indicators to evaluate driver readiness. Furthermore, we created 
%Our framework, used to establish 
%a set of ground truth labels to address the challenge of limited ground truth availability in the context of driver readiness assessment studies.

Our main research contribution can be summarised as follows:
\begin{itemize}
    %\item We investigated the strengths and weakness of the existing models and datasets %concentrate on investigating methodologies 
    %for assessing driver readiness in conditionally automated vehicles.
    \item We introduce an %ground truth framework aimed at assigning numerical values to driver readiness at each moment, 
    extended dataset with new ground truth labels with numerical values for driver readiness, addressing the challenge of limited ground truth availability in the context of driver readiness assessment studies.
    \item We leverage state-of-the-art Convolutional Neural Network (CNN) models in computer vision to enhance the accuracy of feature extraction from eye-tracking data and head pose, establishing them as reliable indicators for driver readiness evaluation.
    %\item We develop a customised Long Short-Term Memory (LSTM) model to analyse the temporal dependencies across the frames.
    \item We employ a Recurrent Neural Network (RNN) architecture that utilises Long Short-Term Memory (LSTM) modules to analyse spatio-temporal dependencies within the video frames recorded from drivers.
\end{itemize}
 
The rest of this paper is organised as follows: Section \ref{sec2} presents the literature review, which examines the existing knowledge on automated driving systems, the importance of NDRTs, the challenges of transitioning between automated and manual driving, and the landscape of vision-based driver monitoring techniques. It also delves into previous research on driver readiness and identifies research gaps. Section \ref{sec3} outlines the dataset used for the study and explains the creation of a ground truth framework to measure driver readiness. In Section \ref{sec4}, the methodology is detailed, covering the extraction of head pose and gaze direction features, the implementation of machine learning models, and the training and validation strategies. Section \ref{sec5} evaluates the performance of the proposed models using various metrics, analyses the impact of different configurations, and presents the obtained results. Finally, in Section \ref{sec6}, the conclusion synthesises the findings, reiterates the research objectives, discusses the implications of the study's outcomes, and suggests potential directions for further research in automated driving and driver readiness assessment.

\section{RELATED WORK}\label{sec2}

Decision-making to pass the driving control to a human driver would need to rely on empirical data regarding drivers' abilities and behaviours in specific situations. For instance, if an analysis of the driver's visual attention indicates that the driver is completely disconnected from the driving task, the request for control transfer could be postponed until their attention is back on driving. Alternatively, if the analysis reveals a challenging driving scenario, the vehicle might execute a low-risk safety manoeuvre to position itself on the roadside \cite {louw2017human}. Thus, it is crucial for a DMS to consistently evaluate the driver's readiness to take-over, which is essential to ensure a safe and timely transfer of control \cite{dogan2022evaluation}. The most accurate depiction of the driver's state can be achieved using physiological-based DMS; however, these systems are excessively intrusive for application in commercial vehicles \cite{deo2019looking}. To address this, the development of DMS has largely shifted towards leveraging computer vision and deep learning technologies \cite{ortega2021real}. 

\subsection{Vision-based driver monitoring}
Driver behaviour analysis based on vision sensors has been a popular topic of research, with a large body of literature focusing on the driver's gaze estimation as a useful cue for estimating the driver's attention \cite{wang2021vision}. By analysing visual cues from the driver's face, these systems can detect signs of drowsiness, distraction, or other cognitive impairments \cite{hayley2021driver}. Initial research relied on assessing the driver's gaze through head pose estimation. In this context, Lee et al. \cite{lee2011real} introduce a vision-based approach to instantly determine the driver's head orientation and gaze direction. This technique remains effective despite changes in facial appearance due to wearing glasses and functions proficiently during both daylight and nighttime scenarios. The method employs facial features and a geometric facial model to gauge the driver's yaw and pitch angles. Furthermore, the approach includes a gaze estimation technique that employs support vector machines (SVM) to accurately calculate both the gaze directions and gaze %zones
regions. In the study by Tawari et al. \cite{tawari2014continuous}, they introduced a distributed camera system that analyses head movements. The main strength of the model is its capability to work reliably and consistently, even when dealing with significant head motions. The system's core function involves tracing facial features and examining their spatial arrangement. This process helps gauge head orientation through a 3-dimensional model.

Recent methods mainly utilise a combination of both head and eye features. Vasli et al. \cite{vasli2016driver} presented a system for estimating a driver's gaze by considering cues from both the head and eye poses. They employed a multi-plane geometrical setting and integrated this with a data-driven learning approach. The researchers evaluated their methods using real-world driving data, encompassing diverse drivers and vehicles, to assess how well the techniques could be applied broadly. Similarly, Fridman et al. \cite{fridman2016owl} investigated the improvement in classifying driver gaze by utilising both head and eye poses instead of relying solely on the head pose. They also explored whether individual-specific gaze patterns were linked to the extent of gaze classification enhancement when including eye pose data. The key insight from their work is metaphorically explained using the concepts of an 'owl' and a 'lizard.' This metaphor illustrates that when the head exhibits substantial movement ('owl'), the addition of eye pose information does not significantly enhance gaze classification. Conversely, when only the eyes move while the head remains still ('lizard'), including eye pose data substantially boosts classification accuracy. The authors also investigated how this accuracy variation relates to different individuals, gaze strategies, and gaze regions.

Recent studies involving CNNs have enabled the accurate prediction of driver gaze areas, which can be applied universally to various drivers and slight deviations in camera positioning. Vora et al. \cite{vora2017generalizing} addressed the challenge of creating a generalised gaze zone estimation system that remains consistent across various subjects, perspectives, and scales. To assess the effectiveness of their system, they gathered extensive real-world driving data and trained their CNN model, achieving high accuracy. Similarly, Lollett et al. \cite{lollett2022single} tackled the challenge of accurately gauging a driver's gaze in real-world scenarios, where factors like facial obstructions and varying distances from the camera can complicate the process. %They used 3D CNN to capture both spatial and temporal driver actions, enhancing accuracy by considering motion-related features from consecutive frames. Their evaluation demonstrated that the 3D CNN model notably outperformed the 2D CNN model, showcasing its efficacy in scenarios with different driver-to-camera distances.

While existing studies have shown promise in accurately predicting driver gaze areas using CNNs and computer vision techniques, only a limited number of studies have focused on the application of gaze zone estimation for evaluating driver readiness in driving scenarios.

\subsection{The driver readiness studies}
In order for the human driver to take back control of the vehicle safely, the driver must be in an appropriate state of readiness prior to the TOR alarm. According to ISO/TR 20195-1 \cite{RN223}, driver readiness refers to the condition of the driver while engaged in automated driving, which impacts their ability to effectively take back control of the vehicle from the system in order to resume manual driving. 

Braunagel et al. \cite{braunagel2017ready} present an advanced driver assistance system that predicts the driver's take-over readiness in conditionally automated vehicles. The system proactively warns the driver if low take-over readiness is anticipated. It determines readiness based on traffic complexity, the driver's secondary task, and gaze direction. An evaluation using a driving simulator with 81 participants demonstrated a 79\% accuracy in predicting take-over readiness. %The system also considers how the take-over intervention type affects predictions. 

Marberger et al. \cite{marberger2018understanding} research focused on human performance during take-over situations and how the driver's state impacts these scenarios. The study established a comprehensive model for the transition process from automated to manual driving. This model defined specific time points and time intervals critical to this transition. The researchers introduced the concept of "Driver Availability," quantitatively measuring the time needed for a safe take-over against the available time frame. %They also presented a conceptual framework outlining potential factors influencing driver availability and practical ways to apply this metric in real-time applications.

Deo and Trivedi \cite{deo2019looking} conducted a study %where they proposed a new approach 
to assess driver readiness in conditionally automated vehicles %. They achieved this 
by analysing information from in-vehicle cameras. Human evaluators examined the camera data and gave their subjective rates for the driver's ability to take-over control. The research introduces the term 'Observable Readiness Index (ORI)', which quantifies the driver's readiness based on these evaluations. Furthermore, the study presents an LSTM model that continuously estimates the driver's ORI by considering various aspects of the driver's state, including gaze, hand position, posture, and foot activity. 

In the research by Arslanyilmaz et al. \cite{arslanyilmaz2020driver}, they explored factors influencing the duration and quality of take-overs during system failures before intersections. These factors encompass secondary task characteristics, traffic density, time buffer before TOR, and the type of TOR warning. The study's primary aim was to investigate how different time buffers (3 and 7 seconds) impact take-over duration and quality in case of system failures near intersections. They also aimed to analyse drivers' eye-tracking behaviour in these time buffers, comparing how drivers focus on information on the dashboard and the road. 

In the study by Kim, J. et al. \cite{kim2022novel}, the focus was on investigating how NDRTs impact a driver's readiness and take-over performance. Through a driving simulator experiment, the researchers assessed driver take-over readiness during the transition from automated to manual driving while participants engaged in various NDRTs. %The findings revealed that NDRTs significantly influenced driver readiness, subsequently affecting their ability to take-over. 
The study established that driver readiness had a negative correlation with take-over time but a positive correlation with the quality of vehicle control.

In the study conducted by Greer et al. \cite{greer2023safe}, the focus was on achieving safe transitions from automated to manual control in vehicles by understanding the driver's situational awareness. The researchers introduced two metrics, the ORI and Take-over Time, which quantify this awareness state. Their model used feature vectors that included hand location, foot activity, and gaze location as input. Additionally, the study introduced two new metrics to assess the quality of control transitions after the take-over event, namely the maximal lateral deviation and velocity deviation.

The current state of research on driver readiness in the context of transitioning from automated to manual driving has taken significant steps forward in understanding the factors influencing this critical process. The literature review highlights the importance of driver engagement in NDRTs, the challenges posed by the transition to manual driving, and the essential role of vision-based DMS in assessing driver readiness. However, certain challenges and research gaps necessitate attention. %Two main challenges stand out: 
First, the lack of comprehensive naturalistic driving datasets capturing a wide range of driver behaviours in conditionally automated vehicles hinders the development of data-driven approaches to map driver activity to take-over readiness effectively. Such datasets would significantly enhance the applicability of research findings. Second, defining a clear and objective ground truth for take-over readiness remains a complex task. As data-driven approaches heavily rely on accurate ground-truth information, the absence of a universally accepted measure for readiness poses a challenge in designing robust models and evaluations. Addressing these gaps would lead to more accurate and reliable methods for assessing driver readiness during the transition from automated to manual driving, ultimately contributing to safer and more efficient automated driving systems.

\section{DATASET AND GROUND TRUTH} \label{sec3}

%In this study, we utilised a video from the Driver Monitoring Dataset (DMD) \cite{ortega2020dmd} as it meets the prerequisites %requirements
%of this study as follows:

In this study, we randomly selected 15 video clips, each spanning approximately 12 seconds, from the S6 category of the Driver Monitoring Dataset (DMD) \cite{ortega2020dmd}.
%which has 10 available videos (averaging 3 minutes in duration, with only male participants). 
This selection meets the following prerequisites of this study:

\begin{itemize}
    \item High-Resolution videos: The dataset includes 1280x720 pixels RGB videos that are zoomed on the driver's face only. % the driver's facial images with a high resolution of , employing an RGB camera. This high-resolution imaging 
    This ensures capturing detailed facial features and expressions which are crucial for assessing driver readiness.
    \item Gaze Estimation Relevance: The scenarios depicted in the dataset finely correspond to the gaze estimation requirements necessary for our study. This alignment between the dataset's content and our research focus enhances the dataset's suitability for our investigation.
\end{itemize}

\begin{figure}
    \centering
    \includegraphics[width=1\linewidth]{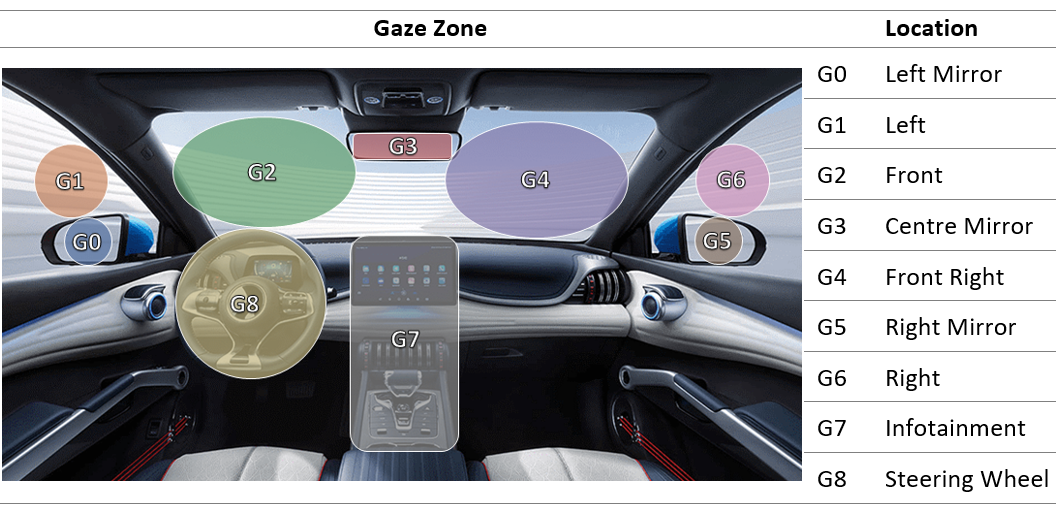}
    %\caption{Classification and positioning of gaze zones within the vehicle cabin \cite{ortega2020dmd}} 
    \caption{Classification and positioning of gaze zones within the vehicle cabin}
    \label{fig: region}
\end{figure}

%The selected video has a length of 3 minutes and 9 seconds, captured at a rate of 30 FPS, resulting in a total of 5682 frames. These frames have been split into 90 sections, each lasting two seconds. To maintain the consistency of these two-second spans, the last 42 frames have been excluded from the dataset.
The selected video clips, captured at a rate of 30 FPS, result in a total of 5,640 frames available for evaluation. These frames have been split into 90 sections, each lasting two seconds.
Furthermore, the chosen videos were recorded in a controlled environment inside %an unmoving 
a real car cabin. Participants were directed to focus their attention on specific regions of interest in the car (regions G0 to G8), as shown in Figure \ref{fig: region}. This deliberate focus on distinct regions facilitates the subsequent estimation of gaze positions, which is crucial for the analysis conducted in our study.

\subsection{Human ratings for driver readiness}
One of the fundamental challenges in driver readiness assessment for take-over control is the absence of publicly available data that includes relevant ground truth information regarding driver readiness. To address the issue of ground truth, we employed a rating method by two human evaluators, similar to the approach taken by studies \cite{li2014predicting} and \cite{deo2019looking}. Human evaluators watch frame feeds and assign a value to the driver readiness based on their head movements and eye gaze. This methodology relies on the ability of experienced drivers to identify and assess driver readiness from video recordings of the driver's face. By analysing the driver's head movements and eye gaze, the evaluators can provide consistent ratings of the driver's readiness levels. A substantial level of agreement among evaluators would indicate a metric for assessing the driver's readiness for take-over that could be determined solely through visual cues \cite{deo2019looking}. The average value offered by the evaluators can serve as the ground truth for training and testing a machine learning algorithm for take-over readiness estimation.

\subsection{Protocol for collecting ratings} \label{Protocol for collecting ratings}
Two individuals holding %valid 
UK driving licenses, and familiar with the functioning of Level 3 (L3) automated driving systems were selected as evaluators. A time span of 2 seconds was utilised to evaluate driver readiness. To achieve this, a sequence of 12 frames, equivalent to 2 seconds at a frame rate of 5 FPS, was presented to each evaluator. Independently, they assigned a readiness score ranging from 1 (lowest driver readiness) to 5 (highest driver readiness) to the driver's state at the culmination of that 2-second interval. The assessment criteria for these readiness levels are outlined in Table \ref{tab: assessment-levels}.

\begin{table*}[!t]%
\centering %
\caption{Driver readiness assessment levels based on head pose and eye direction.\label{tab: assessment-levels}}%
\begin{tabular*}{\textwidth}{@{\extracolsep\fill}cllll@{\extracolsep\fill}}
\toprule
\textbf{Rating} & \textbf{Driver Readiness}  & \textbf{Description} \\
\midrule
1 & Non-attentive  & The driver's head pose and eye direction indicate a complete lack of readiness to take control.\\
2 & Low attentiveness  & The driver's head pose and eye direction suggest minimal awareness and readiness to take control.\\
3 & Partially Ready  & The driver's head pose and eye direction show some awareness, but their readiness is limited.\\
4 & Moderately Ready  & The driver's head pose and eye direction demonstrate moderate readiness to take immediate control.\\
5 & Fully Ready  & The driver's head pose and eye direction indicate full attentiveness and readiness to take control.\\
\bottomrule
\end{tabular*}
\begin{tablenotes}
\end{tablenotes}
\end{table*}

During the rating process, the evaluators could review each segment and update their previous ratings. Although selecting 2-second intervals minimises evaluators' confusion and accelerates the rating process, it leads to the creation of a discrete rating scale. 

\subsection{Readiness index as ground truth} \label{Readiness index as ground truth}
We take the average of the ratings provided by the evaluators to create a single rating score for each 2-second interval. We employ the cubic interpolation method, as described in \cite{waykole2023interpolation}, to extend this discrete rating set to ensure a continuous representation of driver readiness for take-over. This interpolation method is chosen based on its ability to create smooth and accurate curves between data points, capturing the gradual changes in driver readiness over time. While other interpolation methods, such as linear, polynomial, spline, and nearest neighbour, were considered as alternatives, cubic interpolation was selected due to its capacity to handle complex trends, reduce oscillations, and align with the expected gradual changes in human readiness state \cite{deng2014continuous}.

Consequently, for each frame, which corresponds to a time of 1/30 second, an interpolated value is generated for the readiness index using cubic interpolation. This value serves as the foundation for training and evaluating the driver readiness evaluation model, acting as a representative ground truth.

\subsection{Qualitative analysis of readiness ratings}
\begin{figure}
    \centering
    \includegraphics[width=1\linewidth]{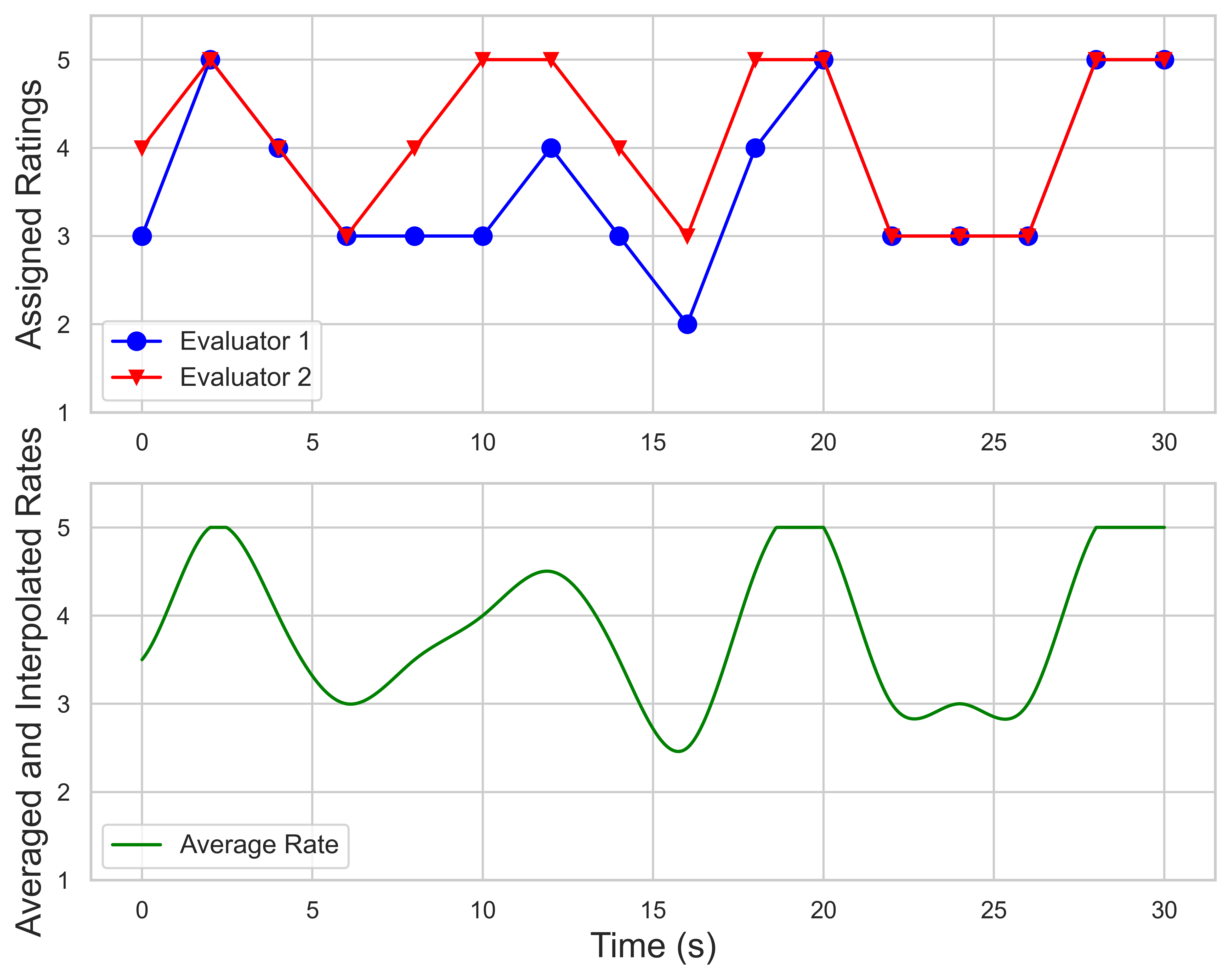}
    \caption{Assigned ratings and derived readiness index for a 30-second time interval.}
    \label{fig: assigned-ratings}
\end{figure}

In Figure \ref{fig: assigned-ratings}, an illustration of a representative evaluation during a 30-second timeframe is displayed. The upper row showcases the ratings designated by individual evaluators. It is evident from the illustration that the ratings display a favourable level of convergence. While there seems to be a lower level of strictness in the rating assignment by evaluator 2, the overall rating trend remains consistent. The lower row, on the other hand, portrays the readiness index in the form of a continuous variable. This index is derived through the process of averaging and cubic interpolation of the provided rates.

\section{METHODOLOGY} \label{sec4}

We aim to evaluate the readiness index continuously based on processing recorded driver videos. This issue is addressed as a regression problem. 
%At each moment, the video recorded from the past two seconds is taken as input to the model, and the output is an estimated measure of driver readiness for take-over.
Drawing inspiration from study \cite{deo2019looking}, at each moment, the video recorded from the past two seconds is taken as input to the model, and the output is an estimated measure of driver readiness for take-over.
CNN-based models are utilised to extract head pose and eye-tracking features from frames related to each two-second interval. Subsequently, these frame-level features are concatenated to represent a frame-by-frame depiction of the driver's state.

\begin{figure*}
    \centering
    \includegraphics[width=0.7\textwidth]{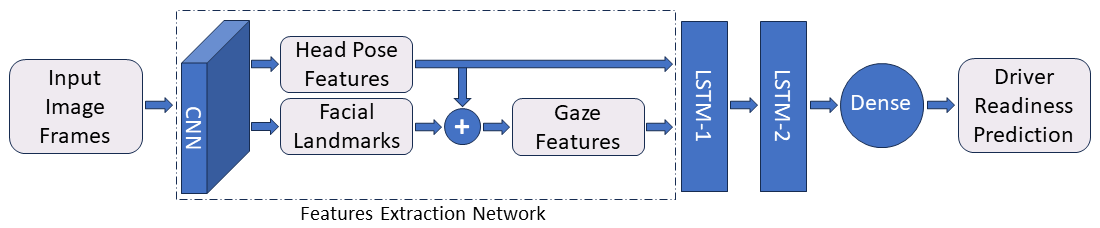}
    %\caption{Deep learning architecture for driver readiness evaluation.}
    \caption{Deep learning architecture for driver readiness evaluation.}
    \label{fig: architecture}
\end{figure*}

While CNNs extract general features from a single frame, the decision-making basis for assessing driver readiness should consider spatio-temporal data via multiple frames. Therefore, we need a model capable of reasoning about temporal dependencies to estimate the readiness index. Given the effectiveness of LSTM networks in modelling long-term temporal dependencies in various sequential modelling tasks within the domain of automated vehicles \cite{bonyani2023dipnet, sing2023, deo2019looking}, it is utilised to estimate the readiness index. As depicted in Figure \ref{fig: architecture}, the CNN component extracts complex spatial features from individual frames. Meanwhile, the LSTM component effectively captures the temporal relationships across a sequence of frames.
 
By employing two consecutive LSTM layers, each contributing its unique ability to comprehend temporal dependencies. The final step is provided by a Dense layer, which integrates the combined insights from the entire architecture to generate a precise numerical value that serves as the predicted driver readiness index.

\subsection{Frame-level feature extraction}
We utilise the SPIGA model presented by Prados-Torreblanca et al. \cite{pradostorreblanca2022shape}, for extracting head pose and eye-tracking features due to its demonstrated higher accuracy compared to other state-of-the-art (SoTA) models. SPIGA stands out as a leading model for head pose estimation on the WFLW dataset \cite{wu2018look}, having achieved the lowest MAE mean (º) among its counterparts. Moreover, in landmark extraction, the SPIGA model is also considered SoTA for face alignment on datasets like 300W \cite{sagonas2016300} and MERL-RAV \cite{kumar2020luvli}. This CNN-based model takes recorded facial frames as input and calculates three head pose-related features: Yaw, Pitch, and Roll, and 98 Facial Landmarks for each frame.
Its CNN architecture excels at extracting relevant features, making it ideal for processing the facial frames in our study. Additionally, SPIGA has been successfully tested in challenging lighting conditions, such as those encountered in in-cabin monitoring scenarios, further boosting its suitability for our research context.
%The rationale for choosing this model is supported by substantial evidence due to its alignment with our task:  convolutional architecture's adeptness at image feature extraction, specialised focus on the head pose and eye landmarks, and its recent contribution to the field of computer vision.

\subsection{Head pose estimation}
In computer vision, head pose estimation primarily refers to measuring the orientation of the driver's head %capacity to determine the angle at which a person's head is positioned 
with respect to the camera or %. A more accurate definition of head pose estimation is the capacity to infer 
the global coordinate system. However, achieving this requires understanding the camera's built-in characteristics to correct the perceptual bias caused by perspective distortion \cite{murphy2008head}. A common approach in head pose estimation is predicting the relative orientation in a 3D environment using Euler yaw, pitch, and roll angles \cite{asperti2023deep}.

\begin{figure}
    \centering
    \includegraphics[width=0.9\linewidth]{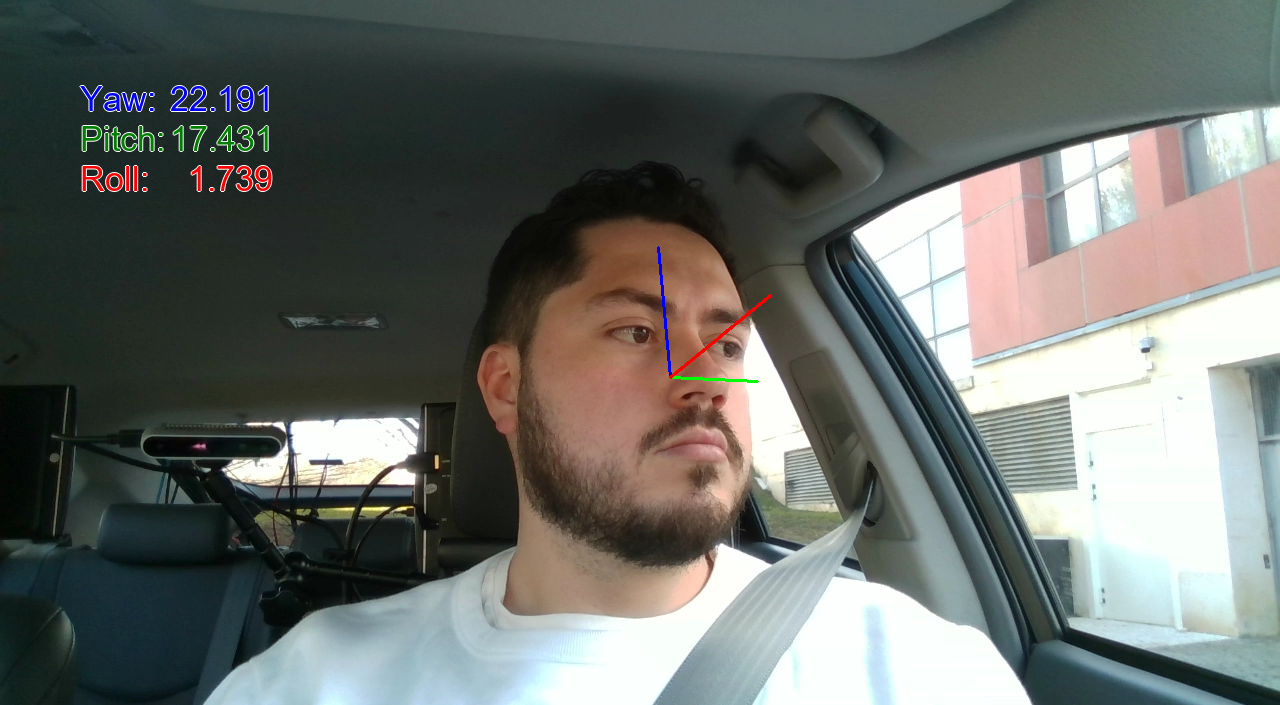}
    \caption{Head pose estimation for a sample frame of DMD dataset using SPIGA model.}
    \label{fig: HPE-SPIGA}
\end{figure}

Although head pose estimation has been a longstanding and extensively explored issue, attaining satisfactory results has only become feasible due to recent advancements in deep learning. Conditions that pose challenges, such as extreme angles, poor lighting, occlusions, and the presence of additional faces within the frame, complicate the task of detecting and estimating head poses for data scientists \cite{asperti2023deep}. This study utilised the SPIGA model, pre-trained on the WFLW dataset \cite{wu2018look}, to extract yaw, pitch, and roll features. Figure \ref{fig: HPE-SPIGA} illustrates the model's output for a frame within the DMD dataset.

\subsection{Eye-tracking data}
Eye-tracking metrics serve as signals of where the visual focus is directed, and this aspect is extremely important for understanding, capturing, and predicting the evolving take-over procedure \cite{zhou2021using}. Automated driving systems should be able to measure and monitor drivers’ situational awareness during the take-over transition period using eye-tracking data with machine learning models \cite{braunagel2017online}. The eye openness ratio provides an understanding of eyelid closure, potentially indicating driver drowsiness or fatigue \cite{singh2022real}. Meanwhile, the horizontal and vertical ratios provide insights into gaze direction and fixation patterns \cite{zemblys2018using}, offering valuable information about the driver's interaction with the environment.

\begin{figure}
    \centering
    \includegraphics[width=0.9\linewidth]{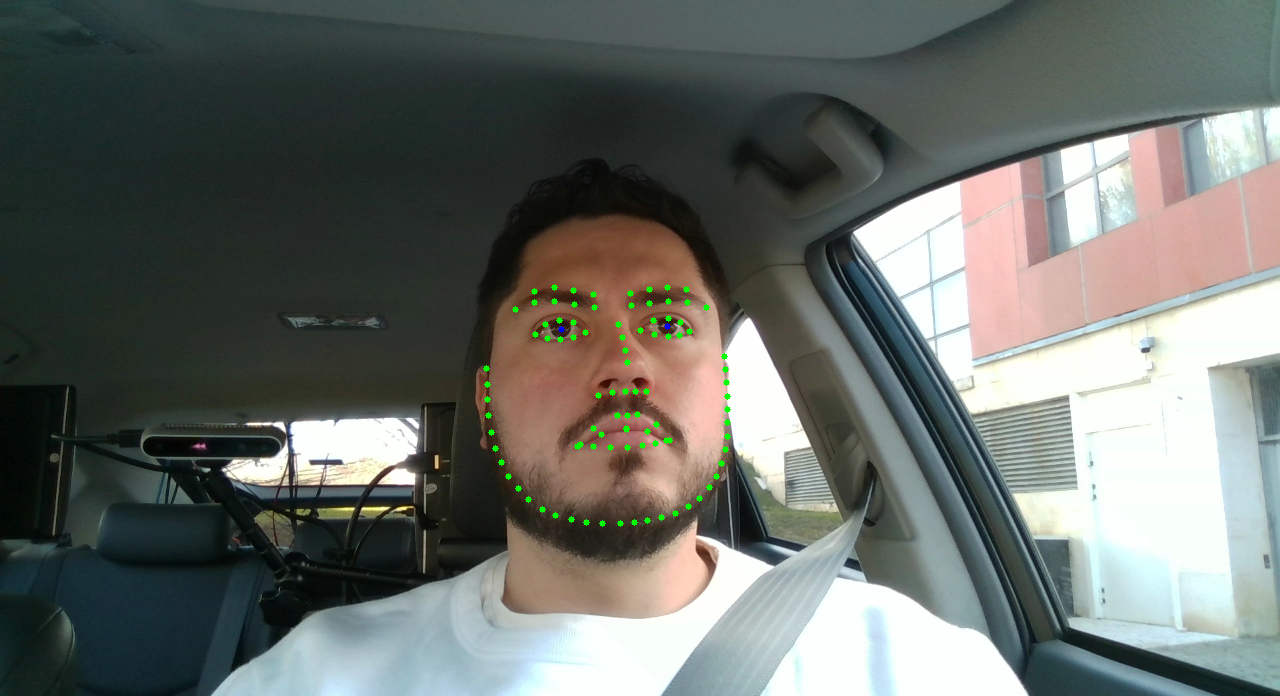}
    \caption{Extracted facial landmarks via SPIGA model for a sample frame.}
    \label{fig: facial-landmarks-SPIGA}
\end{figure}

\begin{figure}
    \centering
    \includegraphics[width=0.9\linewidth]{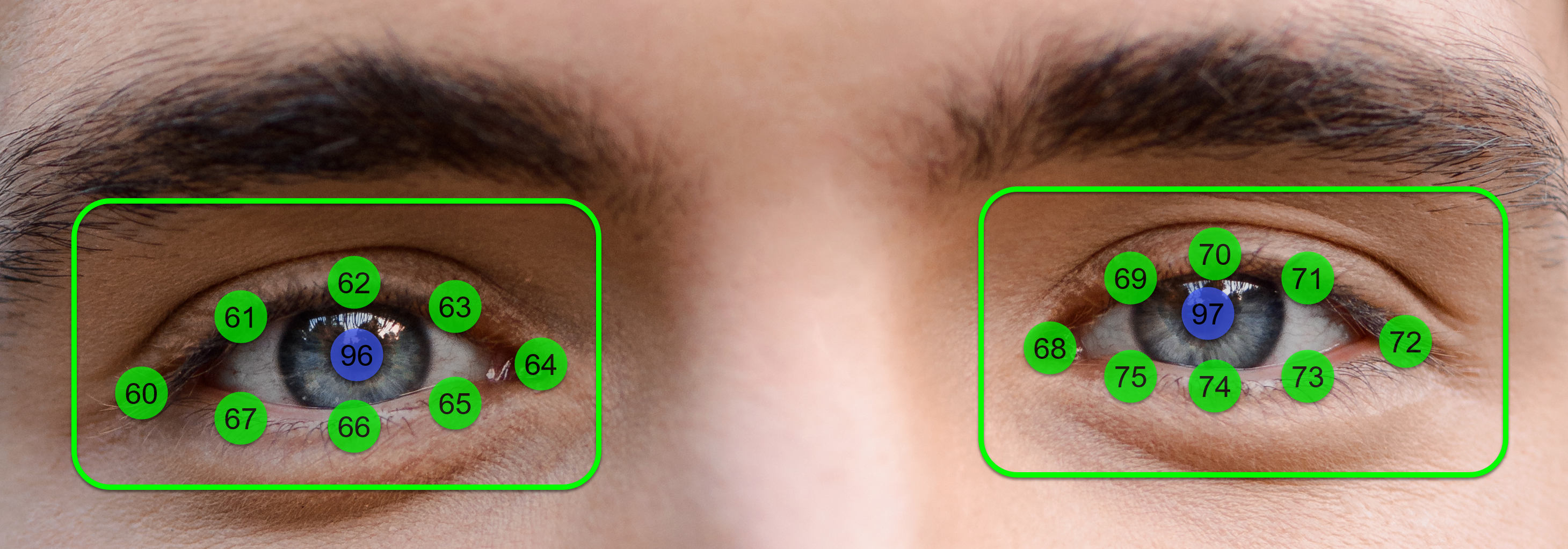}
    \caption{Visual representation of eye and iris landmarks.}
    \label{fig: eye-iris-landmarks}
\end{figure}

To initiate the analysis of eye-tracking data, the process begins with extracting facial landmark points through the utilisation of the SPIGA model \cite{pradostorreblanca2022shape}, as visualised in Figure \ref{fig: facial-landmarks-SPIGA}. Among the 98 extracted facial landmarks, special significance is placed on landmarks 60 to 75, pinpointing the localization of the eyes. Noteworthy are landmarks 96 and 97, crucial for determining the precise positions of the irises. Figure \ref{fig: eye-iris-landmarks} illustrates the spatial arrangement of these key points on the driver's face.

\subsubsection{Eye aspect ratio metric}
Eye Aspect Ratio (EAR) provides an approximate estimation of the eye openness state as the ratio of the vertical distance to the horizontal distance of the eye \cite{maior2020real}. As we employ 98 facial landmarks, the EAR can be defined by the equations below \cite{zhuang2020driver}:

\begin{equation}
    EAR_R = \frac{\|P_{66} - P_{62}\|}{\|P_{64} - P_{60}\|}
\end{equation}

\begin{equation}
    EAR_L = \frac{\|P_{74} - P_{70}\|}{\|P_{72} - P_{68}\|}
\end{equation}

\noindent where $EAR_R$ and $EAR_L$ represent right and left eye aspect ratios, respectively and $P_{dd}$ shows the keypoint number $dd$. EAR values below the threshold of 0.2 indicate that the eyes are closed \cite{dewi2022adjusting}. Furthermore, when there is an extreme head rotation leading one of the eyes to fall outside the camera's field of view, a noticeable difference in the calculated EAR values between the right and left eyes becomes apparent. 

\begin{figure*}
    \centering
    \includegraphics[width=0.9\textwidth]{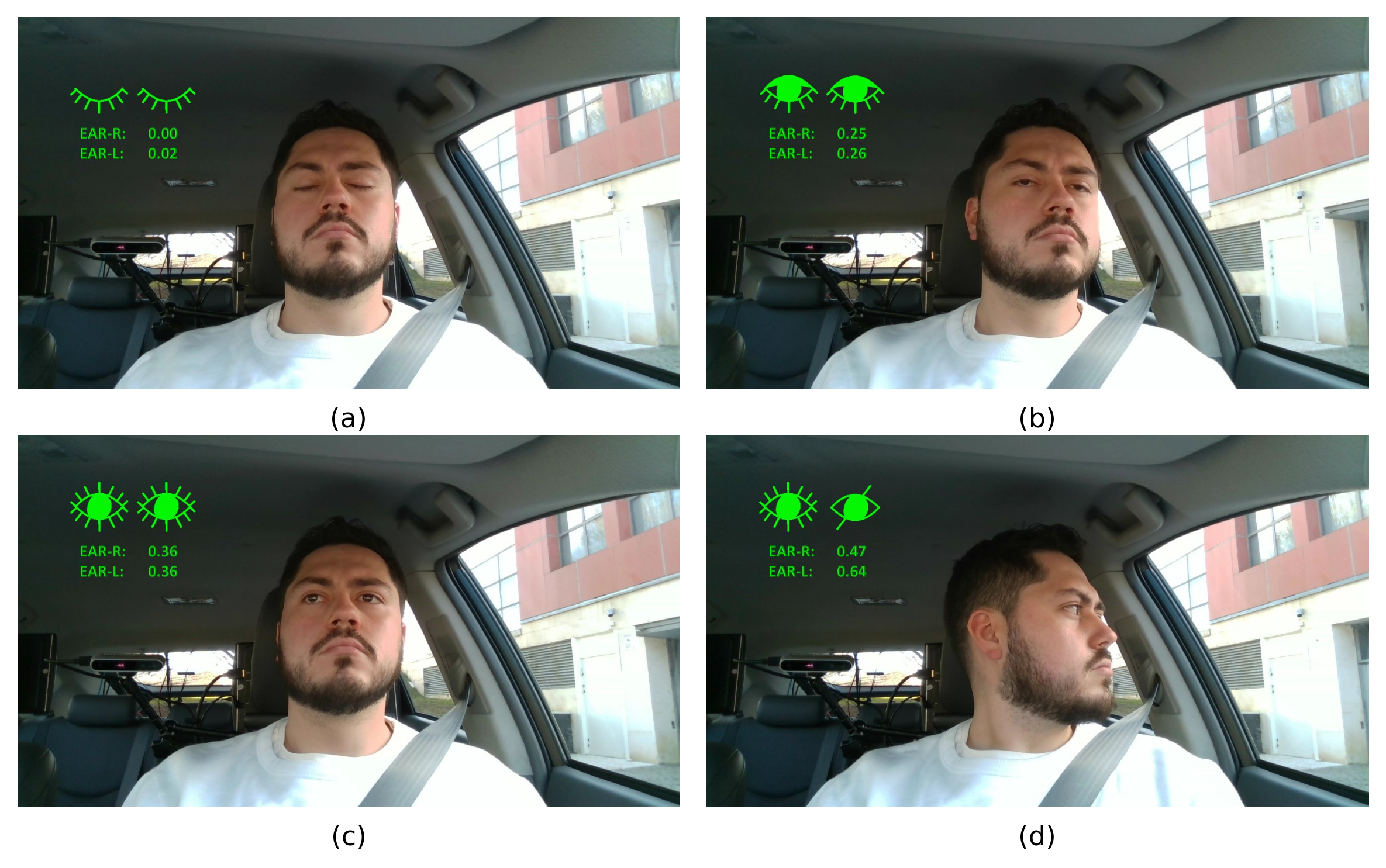}
    \caption{Variation in eye aspect ratio across different eye states and extreme head rotation.}
    \label{fig: eye-aspect-ratio-variations}
\end{figure*}

Figure \ref{fig: eye-aspect-ratio-variations} illustrates the results of EAR computations for four different scenarios: (a) closed eyes, (b) partially open eyes, (c) fully open eyes, and (d) a situation where the left eye falls outside the camera's field of view, due to excessive rotation of the driver's head. 

\subsubsection{Horizontal gaze ratio metric}
The horizontal gaze ratio (HR), ranging from 0.0 to 1.0, indicates horizontal gaze direction. This metric is derived from Ref. \cite{lame2019gazetracking} open-source gaze tracking Python code definitions, utilising the specific landmark x-coordinates ($x_{dd}$). %Additionally, the extent of potential horizontal movement for the pupil is determined by measuring the distance between landmarks 60 and 64. The horizontal gaze ratio is calculated by dividing the right pupil's x offset by the width of the right eye region:

The right eye's horizontal gaze ratio ($HR_R$) is determined by the $x$ offset of the right pupil's landmark 96 from landmark 60, divided by the distance between landmarks 64 and 60:

\begin{equation}
    HR_R = \frac{\|x_{96} - x_{60}\|}{\|x_{64} - x_{60}\|}
\end{equation}

Similarly, $HR_L$ represents the left eye's horizontal gaze ratio, calculated by the $x$ offset of left pupil's landmark 97 from landmark 68, divided by the distance between landmarks 72 and 68:

\begin{equation}
    HR_L = \frac{\|x_{97} - x_{68}\|}{\|x_{72} - x_{68}\|}
\end{equation}

The obtained values reflect the horizontal gaze direction. %A value near 0 indicates a rightward gaze, around 0.5 is straight-ahead, and a tendency towards 1 implies a leftward gaze.
A value within the range [0, 0.33) suggests a rightward gaze, [0.33, 0.67) indicates a straight-ahead gaze, and [0.67, 1.0] implies a leftward gaze tendency.

\begin{figure}
    \centering
    \includegraphics[width=1\linewidth]{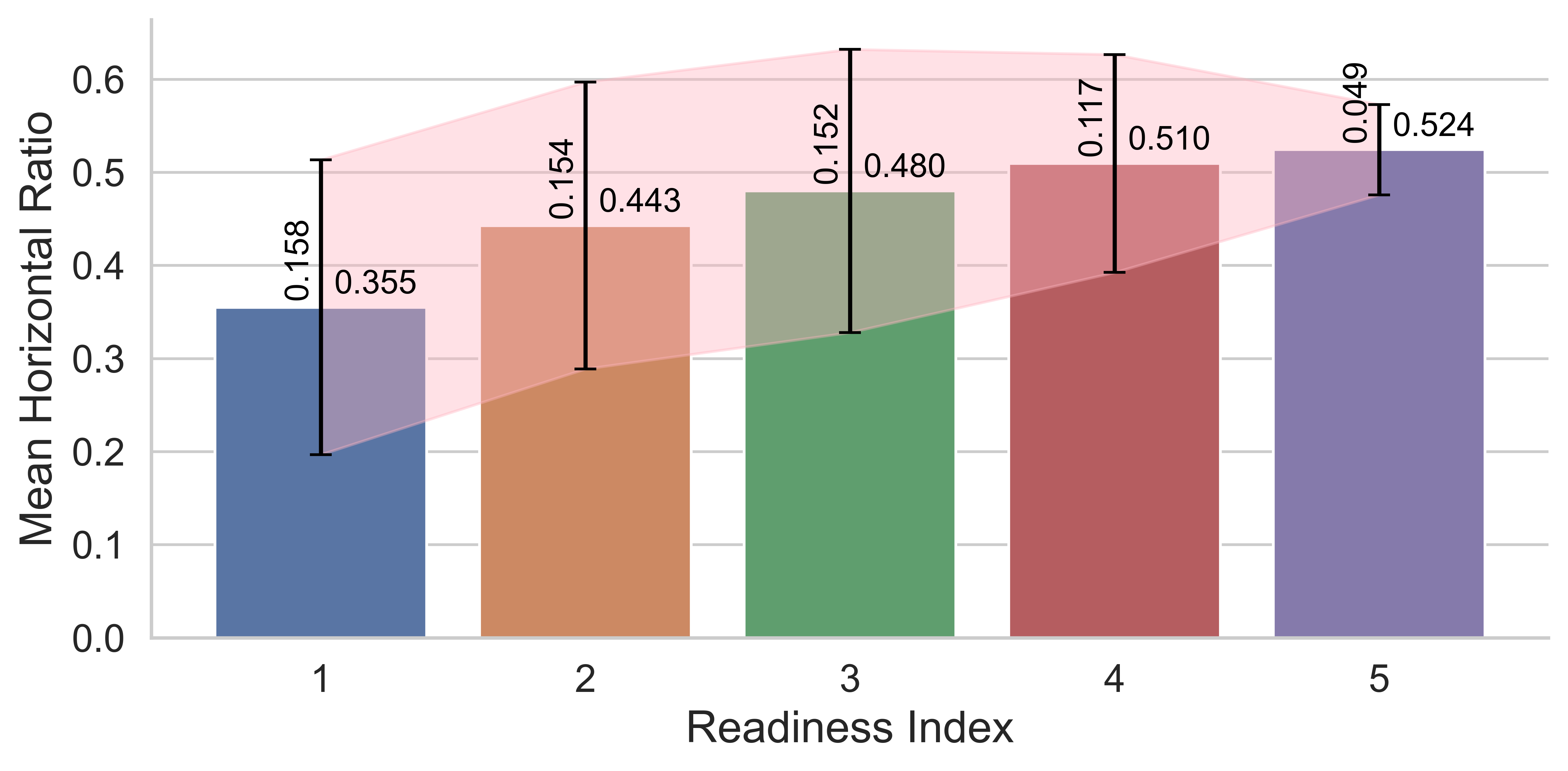}
    \caption{Relationship between driver readiness levels and variability in horizontal gaze ratio.}
    \label{fig: horizontal-gaze-relationship}
\end{figure}

We evaluated the computed values for the horizontal gaze ratio in each of the five driver readiness classes based on the ground truth values generated in Section \ref{Readiness index as ground truth}. Figure \ref{fig: horizontal-gaze-relationship} illustrates the calculated mean and standard deviation for the horizontal gaze ratio in each of these classes. It can be observed that the driver's readiness level increases when the calculated standard deviation decreases. For frames representing the minimum readiness level with a value of 1, we observe the highest standard deviation of 0.158, whereas conversely, for the highest assigned readiness level of 5, the standard deviation reduces more than threefold to 0.049.

This interpretation is straightforward: when a driver possesses a higher level of readiness, their gaze concentration is directed towards a specific forward region, resulting in a slower rate of eye positional changes. This observation aligns with the findings of Ref. \cite{wang2014sensitivity}, which demonstrated that the standard deviation of horizontal gaze position is the most sensitive metric to cognitive focus variations, even though it remains one of the simplest metrics to calculate.

\subsubsection{Vertical gaze ratio metric}
The vertical gaze ratio (VR) is a metric that returns a value between 0.0 and 1.0, indicating the vertical orientation of the gaze. The calculation involves dividing the vertical difference between the pupil's position and a specified reference point (62 for the right eye, 70 for the left eye) by the height of the corresponding eye area \cite{lame2019gazetracking}:

\begin{equation}
    VR_R = \frac{\|y_{96} - y_{62}\|}{\|y_{66} - y_{62}\|}
\end{equation}

\begin{equation}
    VR_L = \frac{\|y_{97} - y_{70}\|}{\|y_{74} - y_{70}\|}
\end{equation}

\noindent where $VR_R$ represents the vertical gaze ratio for the right eye, $VR_L$ for the left eye, and $y_{dd}$ denotes the vertical position of a specific eye landmark $dd$.

In this context, a value in the interval [0, 0.33) %of 0 
corresponds to looking upwards, [0.33, 0.67) %0.5 
indicates a direct gaze, and [0.67, 1] %1 
signifies looking downwards. While the primary use of the vertical gaze ratio is in conjunction with the horizontal gaze ratio for gaze zone classification, values tending towards 1—where the driver's gaze is directed towards the lower portion of their field of view—can suggest a decrease in readiness. This situation may arise when the driver is engaged in an NDRT, such as tuning the radio or using a mobile phone.

\subsection{Gaze zone classification}
Building upon the extracted features related to head pose and eye-tracking data, the driver's gaze area is categorised into nine distinct %detection zones
region of interest (ROI). The selection of zone boundaries follows the definition of the DMD dataset.

While the decision to adopt a nine-%zone
region gaze classification introduces inherent challenges, particularly with gaze regions %zones 
like G0 and G1 due to their close proximity, the availability of gaze region %zone 
annotations for the 2331 frames can provide a robust foundation for training the gaze zone estimation classifier. While we acknowledge that the proposed approach might not compete with the precision of specialised, highly calibrated eye gaze tracker cameras, the introduction of human oversight in the manual review stage serves as a practical strategy to fill the accuracy gap. On this basis, we ensure that a suitable foundation will be established for the study of driver readiness assessment in the following steps.

The main focus is selecting an appropriate classifier based on the extracted features for head pose and eye-tracking data, which can assign the driver's gaze zone to one of these nine ROI in each frame. Two separate scenarios were considered to estimate the driver's gaze %zone 
region using a Random Forest-based classifier. In Case 1, the estimation relied solely on extracted head pose features, while in Case 2, the evaluation was achieved by combining head pose features with additional eye-tracking data. The capability of random forest classifiers in diverse machine learning applications is widely acknowledged, attributed to their skill in directly interpreting acquired parameters and their minimal necessity for hyper-parameter tuning \cite{tawari2014driver}. Following the approach outlined in \cite{tawari2014driver}, we divide the dataset into training and test sets as 80\%–20\%. Specifically, 1864 frames were allocated to training purposes, while 467 were reserved for testing. 

To assess the performance of the trained model, accuracy metric and confusion matrix are employed. Accuracy is a fundamental metric used to evaluate the performance of classification models. It measures the proportion of correctly predicted instances among the total instances in a dataset. It is defined as follows:

\begin{equation}
    Accuracy=\frac{TP+TN}{TP+FN+FP+TN}
\end{equation}

While accuracy provides a general overview of a model's effectiveness, it may not always be sufficient to fully understand its performance, especially in cases where class distributions are imbalanced. This is where a confusion matrix comes into play. The confusion matrix presents a table format that breaks down the model's predictions into four distinct categories: True Positives (TP), True Negatives (TN), False Positives (FP), and False Negatives (FN). These elements provide insights into how well the model distinguishes between classes and where errors occur. By analysing the confusion matrix, it becomes possible to gain a more nuanced assessment of a model’s ability to correctly classify instances across different classes \cite{kim2023study}.
%it becomes possible to calculate additional metrics like precision, recall, and F1-score, which offer a more nuanced assessment of a model's ability to correctly classify instances across different classes \cite{kim2023study}. These metrics are defined as follows:

%\begin{equation}
%    Precision=\frac{TP}{TP+FP}
%\end{equation}

%\begin{equation}
%    Recall=\frac{TP}{TP+FN}
%\end{equation}

%\begin{equation}
%    F1\ score=2\times\frac{Precision\times R e c a l l}{Precision+Recall}
%\end{equation}

\begin{figure*}
    \centering
    \includegraphics[width=1\linewidth]{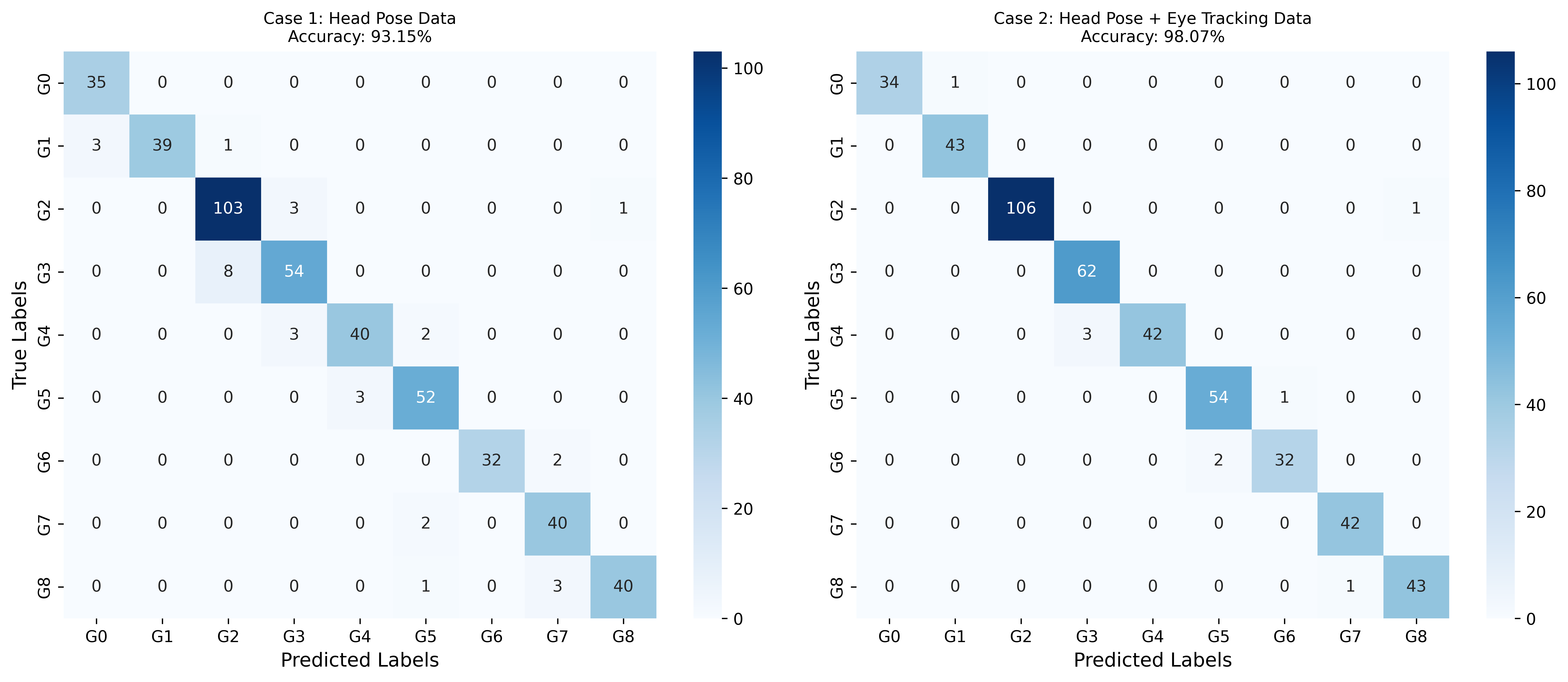}
    \caption{Comparison of confusion matrices and accuracies for gaze zone classification scenarios.}
    \label{fig: gaze-zone-comparison}
\end{figure*}

Figure \ref{fig: gaze-zone-comparison} provides a visual representation of the confusion matrices and the corresponding calculated accuracies for both analysed scenarios. In the case of the classifier that relies solely on head pose features, the accuracy stands at 93.15\%. However, when eye-tracking features are integrated into the existing information, the accuracy experiences a notable enhancement, increasing by approximately 5\% to attain a commendable accuracy of 98.07\%. This improvement underscores the significant contribution of incorporating eye-tracking data into the model's predictive capabilities, resulting in a more refined and accurate gaze %zone
region classification. The augmentation of accuracy by this margin indicates that the integration of eye-tracking data complements and refines the model's decision-making process, enabling it to better distinguish between different ROIs.

Given the accuracy demonstrated by the random forest classifier in gaze zone classification, we proceeded to utilise the model to predict the gaze %zones 
regions for 3309 frames, which had not been previously annotated.

\begin{figure}
    \centering
    \includegraphics[width=1\linewidth]{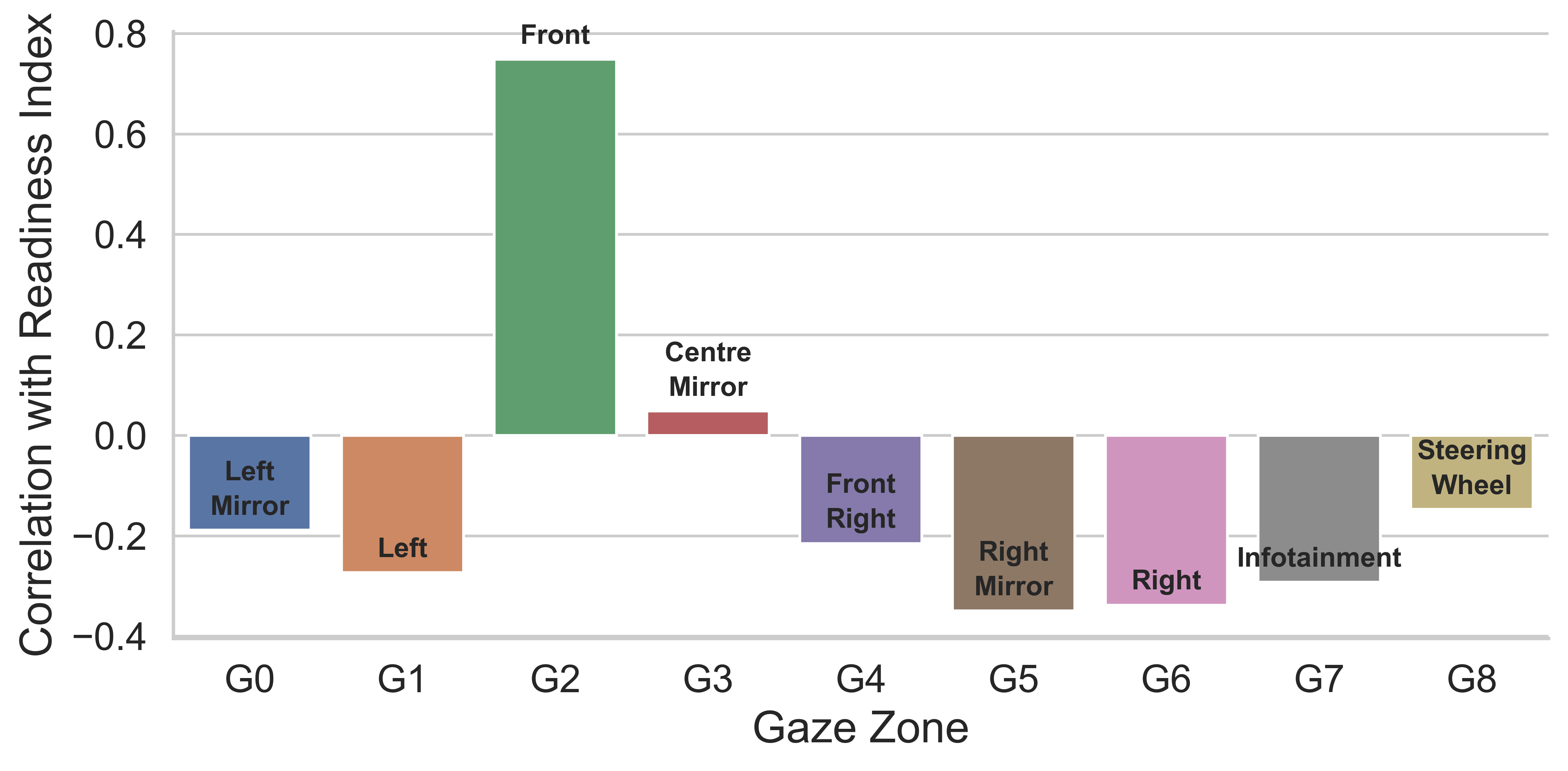}
    \caption{Correlation of driver gaze patterns with readiness levels.}
    \label{fig: correlation}
\end{figure}

Figure \ref{fig: correlation} presents the Pearson correlation coefficient between driver gaze-related features and driver readiness values. The correlation coefficient, a statistical measure, aids in comprehending the strength and direction of a linear relationship between two data sets. Its function is to reveal how closely changes in one variable match up with changes in another variable \cite{lee1988thirteen}. The graph highlights the distinct relationships between different gaze %zones 
regions and the readiness index. Notably, regions associated with forward gaze demonstrate the highest positive correlation with the readiness index. Conversely, gaze regions linked to driver distractions, such as glancing to the left or right, focusing on infotainment systems, or fixating on the steering wheel, exhibit negative correlations with the readiness index. These observations align with the understanding that these actions may divert the driver's attention away from the road, potentially lowering their readiness to respond effectively to unexpected situations.

However, an interesting exception occurs regarding the negative correlations observed with the left and right mirrors. This difference arises from the specific context in which the DMD dataset was generated. In this scenario, drivers have maintained long-lasting and consistent fixation on the mirrors. Evaluators have interpreted this behaviour as indicative of a driver's readiness level reduction. While unexpected, this observation underscores the complexity of interpreting gaze-related features within their unique context.

The random forest classifier categorises the driver's gaze direction by labelling each frame of data based on the driver's visual focus. However, for this data to be effectively utilised by the subsequent machine learning model, which evaluates driver readiness, it needs to be pre-processed. This is where the importance of one-hot encoding comes into play.

In the context of the driver's gaze zone classification, employing one-hot encoding translates to generating a set of binary features, precisely nine in number, each representing a distinct gaze region. The core principle behind one-hot encoding is the creation of new columns, each corresponding to a unique category present in the categorical variable. Subsequently, for every data point, a value of 1 is assigned to the appropriate category-specific column, while all other columns are assigned a value of 0. This transformation enables machine learning models to comprehend categorical distinctions without making any assumptions about the inherent order of categories \cite{seger2018investigation}. Every frame's data is encoded such that one of these features holds a value of 1, indicating the driver's focus on a particular gaze %zone
region, while the remaining features hold values of 0. By adopting this format, the dataset becomes amenable to harnessing the capabilities of neural networks within the domain of machine learning.

\subsection{Time series cross-validation}
With time series data, the concept of cross-validation takes on a new form that adheres to temporal order. Time series carry hidden temporal sequences, predicting future events based on past occurrences. Here, training and validation sets are defined while preserving the temporal order. This approach mimics real-world scenarios where past events guide predictions of future occurrences. Similar to our model, which evaluates the driver’s readiness at each moment based on their head movement and gaze position in previous moments. By embracing temporal dependencies and sequentially changing training and validation windows, this method ensures that future information does not leak into past predictions \cite{bittencourt2020artificial}. It encompasses the nature of time series data and provides a more genuine assessment of model performance. 

\subsection{Proposed LSTM models}
In this research, the dynamic nature of driver readiness assessment demands a model that can capture temporal dependencies within the sequential data. LSTMs excel in this aspect due to their ability to maintain a memory state that can capture information from earlier time steps and propagate it through time. This is crucial for scenarios where the driver's readiness is influenced not only by recent events but also by past interactions with the driver’s state. The integration of both head pose and eye-tracking data as input features aligns well with LSTMs' capacity to handle multiple input streams simultaneously \cite{patil2022comparative}. 

In this study, the effectiveness of two distinct LSTM architectures was explored in the prediction of driver readiness: the Vanilla LSTM \cite{hochreiter1997long} and the Bidirectional LSTM \cite{GRAVES2005602}. For the Vanilla LSTM architecture, the model was constructed with a sequential structure. The architecture commenced with an LSTM layer comprising 64 units, designed to capture temporal dependencies within the input data sequence. The first LSTM layer propagates its outputs to the subsequent layer, preserving the temporal nature of the data. The input shape was defined as window length = 60, accommodating the length of the input sequence and the number of features being considered. Following this, another LSTM layer with 64 units was added, further processing the sequential information. Finally, a Dense layer with a single output unit was included, utilising linear activation to facilitate regression tasks.

For comparison, the Bidirectional LSTM architecture was also structured sequentially. These bidirectional layers inherently captured temporal information from both forward and backward directions, enhancing the model's ability to comprehend complex temporal relationships within the data \cite{deo2019looking}. To ensure a justifiable comparison, we utilise LSTMs with state dimensions of 32 for both the forward and backward segments of the bidirectional LSTM model. Like the Vanilla LSTM architecture, a Dense layer with a single output unit and linear activation was employed for regression tasks.

\subsection{Training and optimisation strategy}
To optimise the model's performance, we employed the Adam optimiser \cite{kingma2014adam}, with a learning rate of 0.001. The learning rate determines the step size taken during each iteration of the optimisation process. A higher learning rate might cause the optimisation process to overshoot the optimal solution, while a lower learning rate might result in slower convergence \cite{sasse2021investigating}. The chosen learning rate of 0.001 balances quick convergence and stable optimisation \cite{zhou2022transfer}, allowing the model to refine its predictions while gradually avoiding drastic oscillations or overshooting.

In the training phase of our models, our primary objective was to minimise the mean absolute error (MAE) between the predicted values and the actual ground truth values within the training dataset. 

\subsection{Validation and test strategy}
We employed a model selection process to determine the most suitable model for final evaluation. This process involved using a validation set that was distinct from the training set. For each training epoch, we calculated the MAE between its predictions and the true values on the validation set. The model that exhibited the lowest MAE on the validation set was chosen as the most optimal one \cite{tekin2020vehicle}. This selection criterion ensured that we identified the model configuration that provided the best predictive performance on unseen data.

Subsequently, the chosen model was evaluated on the test set, which was separate from both the training and validation sets. This set served as an independent measure of the model's performance and generalisation ability. By reporting the results on the test set using the model with the minimum validation set MAE, we aimed to provide an unbiased and reliable assessment of the model's capability to predict the Readiness Index accurately in real-world scenarios.

\section{EVALUATION} \label{sec5}
This section highlights the significance of evaluation metrics, particularly the MAE, in quantifying the alignment between predicted and actual values. We analyse the impact of various parameters on the model's predictions, delving into the influence of cross-validation fold configurations and batch sizes on the model's accuracy. 
Next, we explore the importance of integrating head pose and gaze features within the model architecture for assessing driver readiness. Finally, we present the results obtained from employing different LSTM architectures with distinct feature sets, revealing the model's ability to assess driver readiness in conditionally automated vehicles. Each experiment provides valuable insights into specific aspects contributing to the overall performance and effectiveness of the model.

\subsection{Evaluation of model performance}
Evaluation metrics are essential tools for quantifying the performance of machine learning models. One common metric used for evaluating regression tasks, such as driver readiness evaluation, is the Mean Absolute Error (MAE). 
This metric measures the average absolute difference between the predicted values $\hat{y_i}$ and the actual ground truth values $y_i$, as outlined in Equation \ref{eq:MAE} \cite{sammut2011encyclopedia}.

\begin{equation} \label{eq:MAE}
    MAE=\frac{1}{m}\sum_{i=1}^{m}\left|{\hat{y}}_i-y_i\right|
\end{equation}

MAE provides a straightforward and interpretable measure of how well the model's predictions align with the actual driver readiness status. A lower MAE indicates that the model's predictions are, on average, closer to the actual values, implying better accuracy.

\subsection{N-fold configuration}
Since ensuring consistent test set sizes across all splits was a key consideration in our analysis, the maximum number of folds was constrained by the Ref. \cite{pedregosa2011scikit} formula, and the split of the data was ultimately constrained to a maximum of 5 folds:

\begin{multline}
\text{Validation Size} + \text{Test Size} = \frac{\text{Number of samples}}{\text{Number of splits} + 1} \\
\Rightarrow (400 + 60) + (400 + 60) = \frac{5640}{\text{Number of splits} + 1} \\
\Rightarrow \text{Number of splits} = 5.13
\end{multline}

\begin{figure}
    \centering
    \includegraphics[width=1\linewidth]{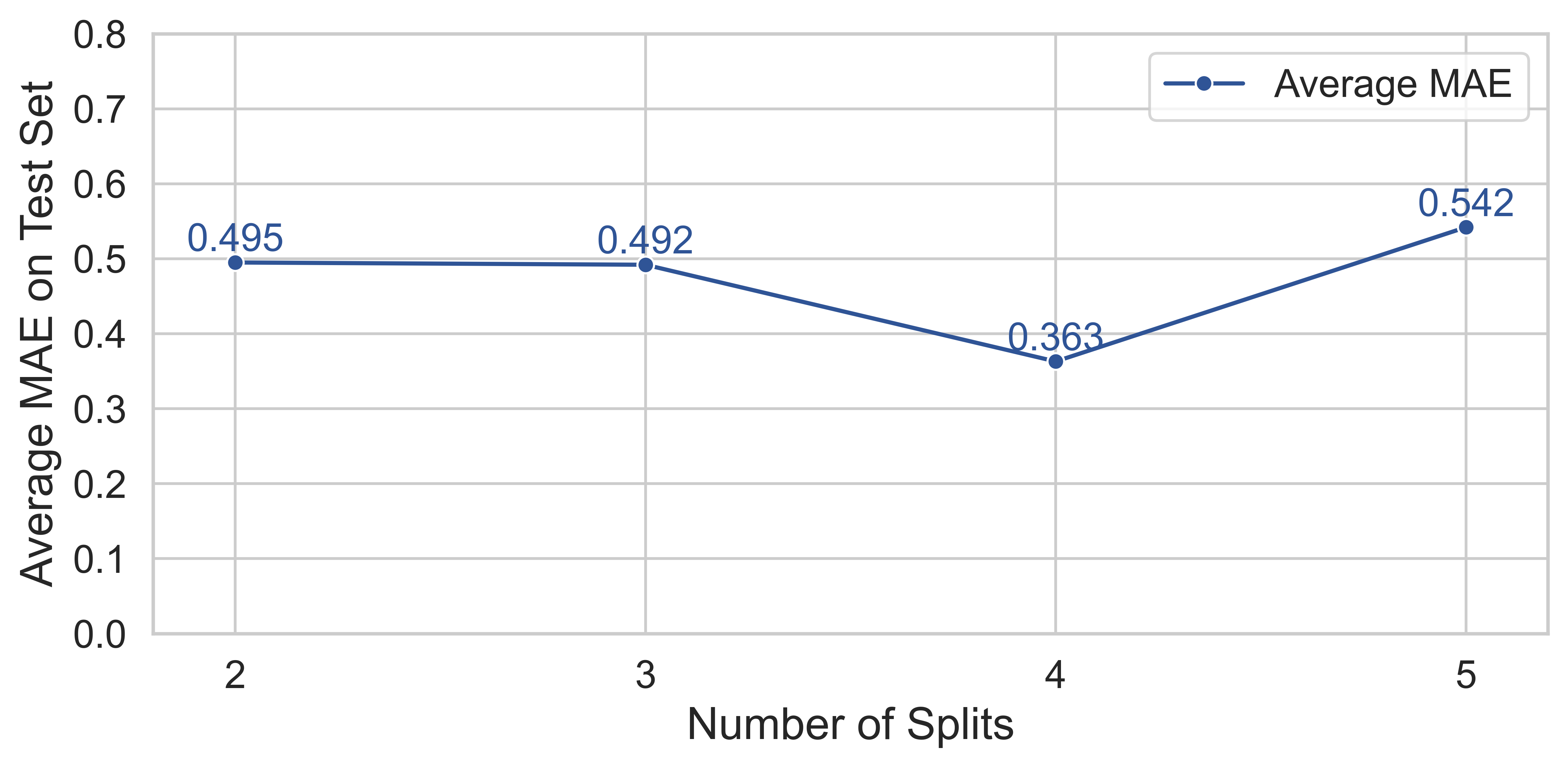}
    \caption{Impact of fold configuration on model performance.}
    \label{fig: fold-configuration-impact}
\end{figure}

To investigate the impact of selecting an optimal number of folds on the performance of the model, with both head pose and gaze features, a thorough analysis was conducted. The results of this analysis are illustrated in Figure \ref{fig: fold-configuration-impact} and provide insight into the relationship between the number of folds and the average MAE of the model.

Interestingly, the results show a trend of decreasing average MAE as the number of folds increases from 2 to 4, reaching its lowest value of 0.363 at four-folds. This suggests that, in this particular context, a higher number of folds enhances the model's generalisation capabilities and predictive accuracy. However, it is worth mentioning that when the number of folds is increased to 5, there is a noticeable rise in the average MAE to 0.542.

\subsection{Optimal batch size selection}
Batch size is a key hyper-parameter which refers to the number of training examples utilised in a single iteration of the model's learning process. In this study, we evaluated different batch sizes (from 1 to 6) to understand their impact on model performance and training efficiency.

\begin{figure}
    \centering
    \includegraphics[width=1\linewidth]{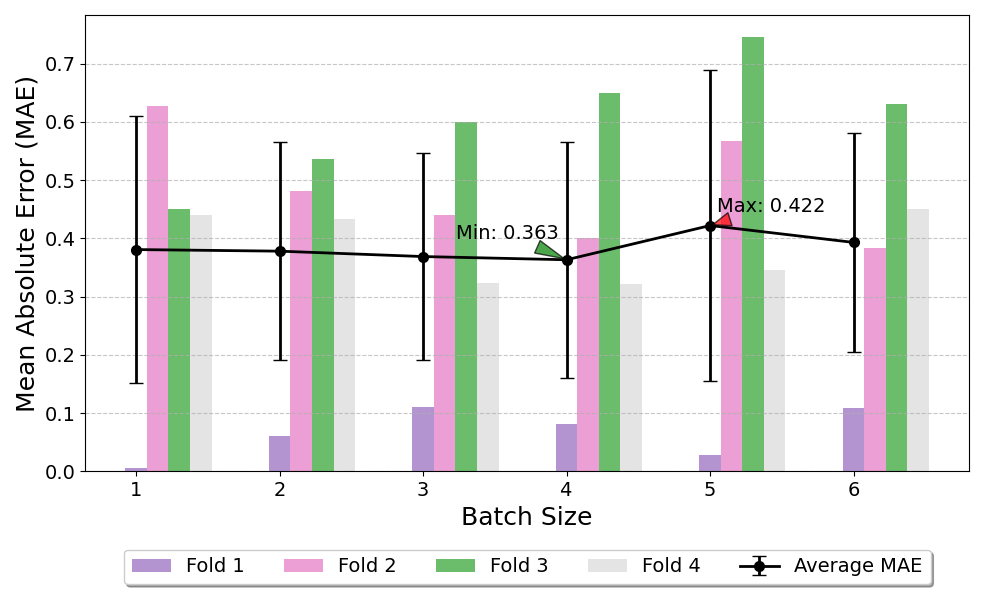}
    \caption{Impact of batch size on model performance.}
    \label{fig: batch-size-impact}
\end{figure}

Upon analysing the results, as shown in Figure \ref{fig: batch-size-impact}, it is evident that different batch sizes have led to varying levels of model performance. Batch sizes 1 and 2 demonstrated low MAE values for certain folds, suggesting potential over-fitting to the training data. Moreover, the high standard deviation of MAE values across folds for these batch sizes indicated inconsistency in performance. Batch sizes 5 and 6, on the other hand, resulted in higher average MAE values compared to batch sizes 2-4, indicating reduced model accuracy. Despite batch size 6 having a low standard deviation, its relatively high average MAE suggests that the model might not capture the underlying patterns effectively.

Considering the trade-off between training efficiency and model performance, batch sizes 3 and 4 emerge as promising options. Batch size 3 produced an average MAE of 0.369 and a relatively low standard deviation of 0.178. Likewise, batch size 4 achieved an average MAE of 0.362 and a standard deviation of 0.209. Considering larger batch sizes can improve training efficiency due to hardware optimisation and parallelism \cite{venkataramanaiah2020fpga}, batch size 4 emerges as the most suitable choice. 

\subsection{Results}
The obtained results, as presented in Table \ref{tab: mae-comparison}, reveal several key insights. Both LSTM architectures exhibited similar predictive capabilities when using only the head features, resulting in MAE values of approximately 0.935 for Vanilla LSTM and 0.934 for Bidirectional LSTM. Moreover, when focusing solely on the gaze features, the Bidirectional LSTM model showed superior performance with an MAE of 0.385, outperforming the Vanilla LSTM's MAE of 0.441. 

\begin{table}[!t]
\caption{MAE calculated for predicted readiness index values in comparison to the ground truth.}
\label{tab: mae-comparison}
\centering
\begin{tabular}{ccccc}
\toprule
\multicolumn{2}{c}{\textbf{Features Used}} & \textbf{Vanilla LSTM} & \textbf{Bidirectional LSTM} \\
\cmidrule{1-2}
Head & Gaze & & \\
\midrule
$\checkmark$ & & 0.935 & 0.934 \\
& $\checkmark$ & 0.441 & 0.385 \\
$\checkmark$ & $\checkmark$ & \textbf{0.375} & \textbf{0.363} \\
\bottomrule
\end{tabular}
\end{table}

Significantly, the integration of both head and gaze features enhanced prediction accuracy. The Bidirectional LSTM model, when considering these combined features set, demonstrated the most robust performance, achieving an MAE of 0.363 compared to the Vanilla LSTM's MAE of 0.375. Consequently, these findings highlight the importance of incorporating both head pose and eye-tracking data, indicating that a bidirectional architecture further improves the model's predictive capacity to assess driver readiness in conditionally automated vehicles.

\begin{figure}
    \centering
    \includegraphics[width=1\linewidth]{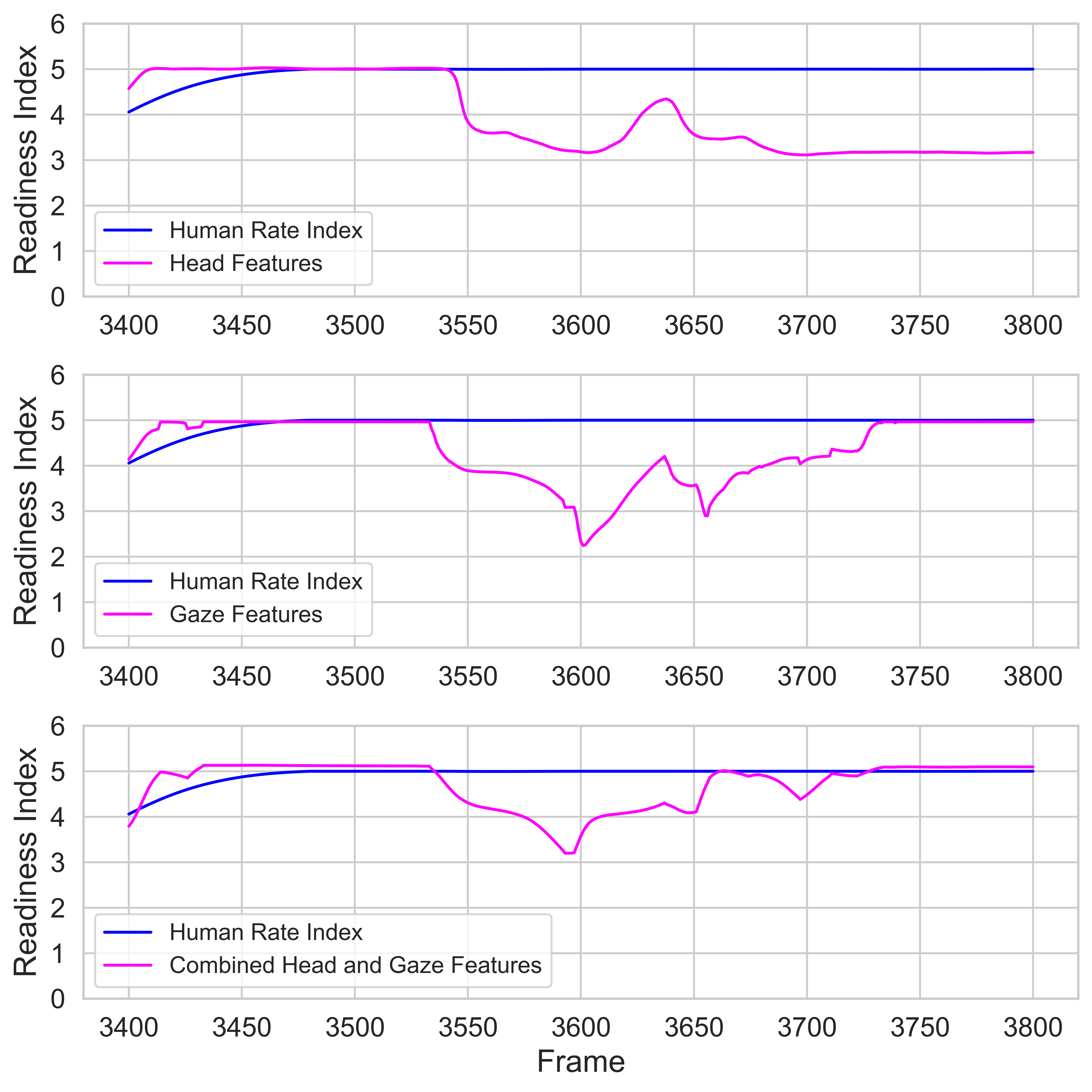}
    \caption{Evolution of model performance through feature integration.}
    \label{fig: model-performance-evolution}
\end{figure}

To provide a qualitative assessment of the model's performance when combining features, an example of data analysis utilising the Bidirectional LSTM model on the test set of the second fold is presented in Figure \ref{fig: model-performance-evolution}. This analysis aims to show how the model's predictions evolve when utilising different sets of features. The top row shows predicted ratings based exclusively on head pose features, whereas the middle row presents ratings solely derived from gaze zone features. The bottom row of the graph displays the model's performance when both sets of features are combined. Upon observation, it becomes evident that the model's performance is notably enhanced when these feature sets are combined.

Specifically focusing on the range of frames spanning from 3530 to 3740, corresponding to 7 seconds of the recorded video, successive changes in the driver's head movement and gaze direction can be observed during this period. Despite these variations, due to the driver's attention to the driving environment remains consistent. The situational awareness has led evaluators to assign a high level of readiness to this interval. However, it is worth noting that while the model does recognise these variations in head pose and gaze direction, its estimation of readiness slightly differs from the evaluators' high rating. This difference might be due to the model's sensitivity to small changes that might not necessarily reflect a reduction in readiness but rather a reaction to the feature change.

Nevertheless, the key point is that the model's performance improves when both head pose and eye-tracking data features are combined. The synergy between these two sets of features allows the model to better capture and understand the complex interplay between the driver's head movements, gaze direction, and overall readiness level.

\begin{figure}
    \centering
    \includegraphics[width=1\linewidth]{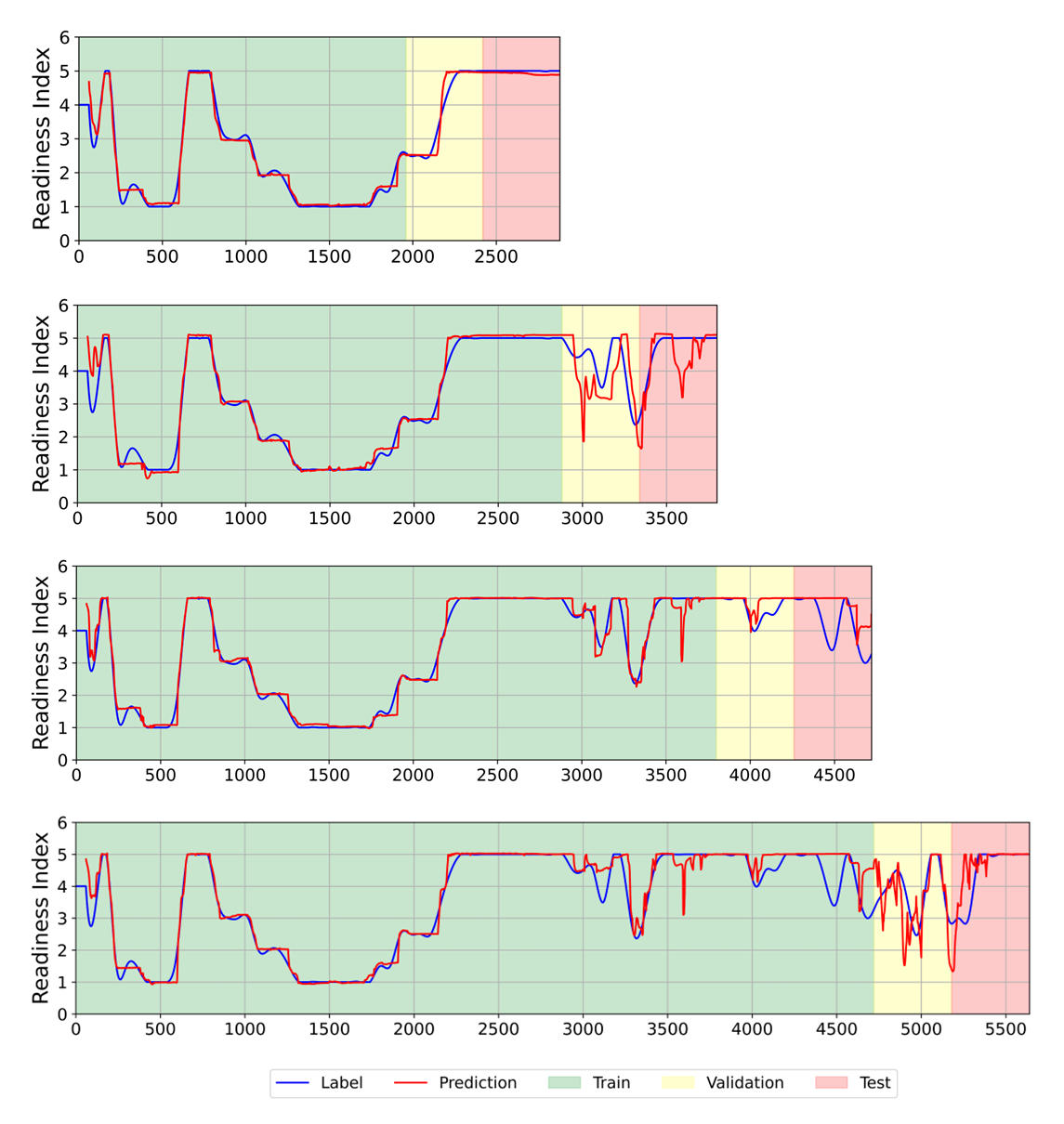}
    \caption{Visualisation of model outputs in different cross-validation folds.}
    \label{fig: model-cross-validation}
\end{figure}

Figure \ref{fig: model-cross-validation} provides a visual representation of the output from the best model in each cross-validation iteration.%, building upon the information introduced in Section 5.2. 
The graphs are colour-coded to aid in understanding the model's learning process. The goal is to find optimal hyper-parameters as the model progresses through different epochs in the training set, represented by the green colour in the graphs. The model's output is assessed using the validation set, denoted by the yellow sections in the figures throughout each epoch. The model's architecture is designed to select the model with the lowest MAE on the validation set as the best model. Once the best model is identified, it is subjected to testing on the red section, which represents a previously unseen test set.

The MAE results attained from the test set are averaged, and this outcome contributes to the overall assessment of prediction performance. In the case of the best model, which employs a Bidirectional LSTM architecture and utilises both head pose and gaze zone features, the achieved accuracy is recorded as 0.363.

It is essential to acknowledge that the presence of noise in the output from the test section is evident. This noise can be attributed to several factors. First, the dataset has limited frames, which could affect the model's ability to generalise effectively. Additionally, an imbalance in driver behaviour across different dataset segments might contribute to the observed noise. For example, more straightforward scenarios at the beginning of the video and more complex situations with rapid variations in head and eye positions toward the video's end can challenge the model's performance. Despite these challenges, the model demonstrates a reasonable level of performance in its predictions when compared to the assigned readiness index values.

\section{CONCLUSION} \label{sec6}

%With the ongoing advancement of automated driving technology, the responsibilities of the driver are set to transform. In vehicles with conditional automation, the driver will retain the duty of monitoring the driving environment and assuming control of the vehicle should the automated system fails or is unable to handle a situation. Consequently, the driver must remain consistently attentive and prepared to intervene whenever necessary. This means that the driver needs to be constantly aware and ready to take action at any time. In this study, we initiated a comprehensive analysis to evaluate driver readiness using a combination of head pose and eye-tracking data. Through a systematically designed approach involving dataset selection, methodology development, and evaluation, valuable insights regarding the effectiveness of predictive models in evaluating driver readiness were obtained.

Efficient feature extraction and integration for driver readiness evaluation was one of the objectives of this study. By combining head pose and eye-tracking features, this model demonstrated significant capabilities in interpreting complex relations between the driver's physical head orientation and direction of attention and focus. LSTMs were selected due to their effective temporal pattern recognition. Bidirectional LSTM architecture, more specifically when initialised with this feature combination, performed optimally, achieving a substantial mean absolute error (MAE) of 0.363. This emphasises that a holistic perspective on driver behaviour provides a more accurate representation of their readiness level. %Additionally, the modular model architecture allows for the addition or replacement of features.

%This research journey was not without its challenges. 
A significant challenge in this research was the absence of a comprehensive driving dataset that encompasses a wide range of driver behaviours in automated vehicles. Additionally, the lack of a clear ground truth for readiness assessment posed significant obstacles. To address these challenges, we created a ground truth set in order to accurately evaluate our model and also for the benefit of the research community in future studies.  

This was done based on the qualitative observations and ratings of multiple human evaluators. 
%framework based on evaluators' observations was used to create a ground truth. 
The model's demonstrated performance at an acceptable level suggests its potential for real-world Level 3 automated driving applications. As technology progresses and data accessibility improves, these obstacles can be reduced, leading to even greater precision and resilience in predictive models.

%This research carries significant practical consequences. It is crucial to precisely assess the readiness of drivers in automated vehicles to guarantee the safety and effectiveness of such vehicles. As automation advances from level 2 to level 3, automated vehicles become more advanced, underscoring the importance of confirming the driver's readiness and ability to regain control when needed. This study presents an approach that can be employed to create and implement a system for monitoring drivers. This system not only encompasses intricate technological elements but also tackles the difficulties associated with maintaining safety during the transition from automated to manual driving.

%The model includes algorithms, data integration, evaluation metrics, and real-world concepts. This research has made it easier to explore and improve the model in the future. Future research can expand the scope of the model to include other ways to generate ground truth data. This will improve the model's ability to determine driver readiness. 
The modular model architecture makes it easy to integrate new driver-specific features, such as hand status, body posture, and steering wheel activity, into the model for future research. By combining these different features, the accuracy and predictive power of the model can be improved. This will allow for a more comprehensive assessment of driver readiness in conditionally automated vehicles. The model's adaptability and real-world applicability will also improve, making it more useful for ensuring a safe and reliable automated driving experience. 

%\backmatter
\bmsection*{Author contributions}
%Copy from DIPNet!
Mostafa Kazemi: Conceptualisation, formal analysis, investigation, methodology, software, visualisation, writing the original draft. Mahdi Rezaei: Methodology, resources, supervision, validation, visualisation, writing the original draft, review and editing. Mohsen Azarmi: Software, writing, review and editing.

%\bmsection*{Acknowledgements}
%The authors would like to thank all partners within the Hi-Drive project for their cooperation and valuable contribution. This research has received funding from the European Union’s Horizon 2020 research and innovation programme, under grant Agreement No 101006664. The article reflects only the author’s view and neither the European Commission nor CINEA is responsible for any use that may be made of the information this document contains.

\bmsection*{Conflict of interest}
The authors declare no potential conflicts of interest.

\bmsection*{Data availability}
The data that support the findings of this study are openly available in the DMD repository at \href{https://dmd.vicomtech.org/}{https://dmd.vicomtech.org/}.

\bmsection*{ORCID}
%\noindent \textit{Mostafa Kazemi} \href{https://orcid.org/0000-0002-4150-7989} {https://orcid.org/0000-0002-4150-7989}

\textit{Mahdi Rezaei} \href{https://orcid.org/0000-0003-3892-421X} {https://orcid.org/0000-0003-3892-421X}

%\noindent \textit{Mohsen Azarmi} \href{https://orcid.org/0000-0003-0737-9204} {https://orcid.org/0000-0003-0737-9204}

%\bibliography{wileyNJDv5-APS}
\bibliography{wileyNJDv5-APS}
\vspace*{12pt}

%\nocite{*}% Show all bib entries - both cited and uncited; comment this line to view only cited bib entries;

\end{document}